# An Action Recognition network for specific target based on rMC and RPN

Mingjie Li[1], Youqian Feng[1], Zhonghai Yin[1], Cheng Zhou[1*],

Fanghao Dong[1], Yuan Lin[1], Yuhao Dong [1]

[1]Air Force Engineering University, Foundation Department, Xi'an, China

*Corresponding author's e-mail: zhou417461659@163.com

**Abstract.** The traditional methods of action recognition are not specific for the operator, thus results are easy to be disturbed when other actions are operated in videos. The network based on mixed convolutional resnet and RPN is proposed in this paper. The rMC is tested in the data set of UCF-101 to compare with the method of R3D. The result shows that its correct rate reaches 71.07%. Meanwhile, the action recognition network is tested in our gesture and body posture data sets for specific target. The simulation achieves a good performance in which the running speed reaches 200 FPS. Finally, our model is improved by introducing the regression block and performs better, which shows the great potential of this model.

## 1. Introduction

In recent years, the technology of deep learning has been widely used in all walks of life. Especially in the field of target recognition and analysis of behavior and action, several networks based on deep learning have achieved excellent performance in improving the accuracy of target location and action recognition.

In the field of target recognition, the mainstream algorithms include the series of RCNN[1][2][3], YOLO[4][5][6], SSD[7] and R-FCN[8], etc. Meanwhile, all the networks with the highest performance (for instance Faster-RCNN , yolo-v3 and SSD) draw on the basic structure of RPN(region proposal network) which is suitable for single target detection.

In the field of action recognition, two stream networks[9][10][11] have access to good performance through training the spatial and temporal features, separately. At the meanwhile, the series of C3D networks[12][13][14] performs well at extremely fast detection speeds.

However, these methods are useful for action videos in which there is no specific target. Wrong detection will occur if there exists other actions in the video. So it is important to realize action recognition for specific target. This is a composite task of target detection and action recognition.

But as we know, the task of target detection is used in images, while action recognition is used in videos. 2Dcnn is the basic structure of the former while 3Dcnn is used more by the latter. Thus, it will improve speed and simplify structure to realize the fusion of 2D and 3D operation. In the Pseudo-3D network[15] , a 3Dresnet decoupling mechanism is proposed in which 3Dresnet is replaced by a 2Drennet and 3DCNN in the similar manner of depth wise separable convolutions[16]. Furthermore, the concept of mixed convolution and R(2+1)D[17] is also proposed recently. According to the concept of mixed convolution, a mixed convolution resnet-50 is designed firstly in the paper. The model is tested in UCF-101 and compared to R3D. The result shows the structure is effective to get spatial and temporal information. Further, a network which can realize the target detection and action recognition of target according to the fusion of R2D[18] and rMC(Mixed convolutional resnet) is proposed. The model is tested in gesture and body posture data sets which is manufactured by ourselves. In terms of gesture data , the AP of detection reaches 44.7% and the accuracy of action recognition reaches 65.3%. Meanwhile, the former reaches 44.47% when the latter reaches 64.8% in the body posture data. The running speed reaches 200 Fps. Further the model is improved by introducing the regression of proposal boxes. The AP reaches 49.7% and the accuracy reaches 70.7% of gesture data. Its running speed also reaches 117 FPS. A great potential for improvement of our model is shown.

## 2. Related works

*2.1. RPN*

RPN is first proposed by He[19] in Faster-RCNN. It is used to replace the complex select-search process of Fast-RCNN. Through training of neural network structure, the proposal box is recommended to realize the classification and single-target detection. On the basis of this structure, multi-objective detection can be carried out by multi-Objective classification and regression. In this paper, the highest confidence score box of RPN is the input of crop pool layer to crop feature maps to realize action recognition later. The structure of RPN is shown in Fig.1

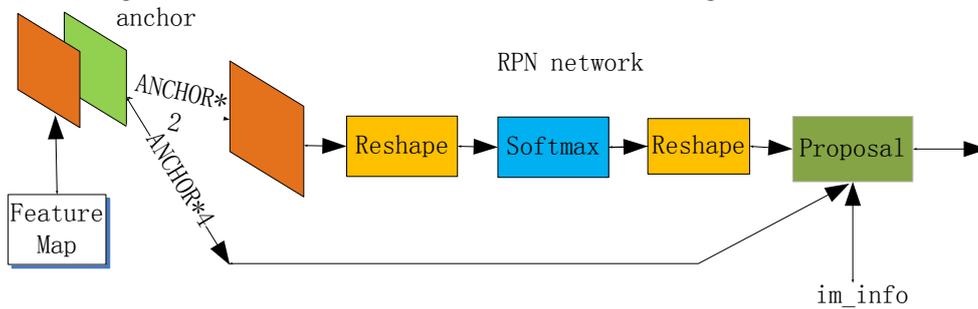

**Fig.1**. Structure of RPN

Fundamentals of RPN is widely used in the algorithms with the best performance (SSD, Yolo-v3,Faster-RCNN, Mask-RCNN[19], etc).

In all the algorithms, it connects to various feature maps to proposal the region of interests. It has a strong mobility to improve effect of target detection and realize multi-target recognition. So it is a front end to proposal a rough region of target in our model to crop the valid region latter.

*2.2. C3D*

C3D is proposed by Tran[17] and used for fast video processing for action recognition. 3D convolution, 3D Pooling and other operations are proposed in C3D. In the 3D operation, video length is treated as channel in 2D convolution. Through this processing, the composite of temporal and spatial features can be realized, and the length of video does not change after 3D convolution (using padding format). The structure of 3D convolution and 3D pooling is shown in Fig.2.

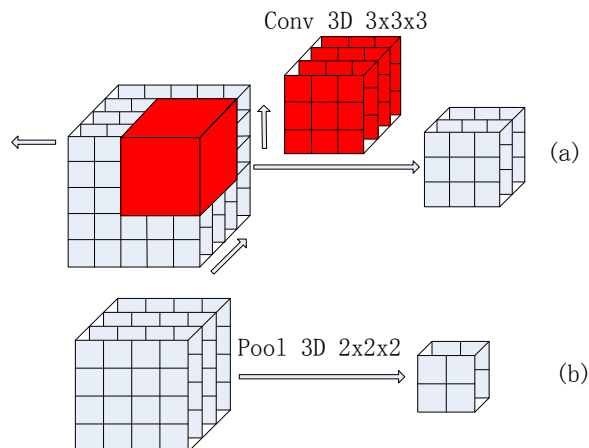

**Fig.2.** The structure of 3D convolution and pooling. (a) is 3Dconv and (b) is 3D pooling.

The traditional C3D has excellent performance in action recognition, motion similarity labeling and other fields. Under the condition of processing 16 frame images at a time, the speed of detection has reached 672 FPS. C3D also has strong mobility, and can be used to be the front-end network of many complex networks. In this paper, the C3D method is used in our mixed convolution resent. It processes the video after the feature maps of target extracted.

*2.3. R2D and R3D*

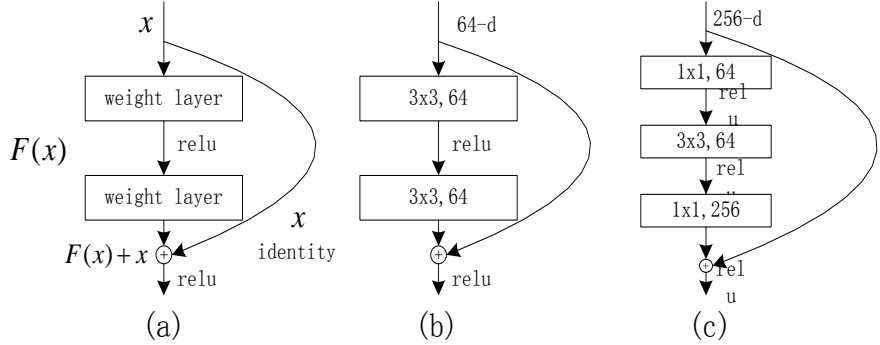

**Fig.3**. The structure of residual block. (a) is the principle of residual block. (b) is the basic structure of Resnet-34. (c) is the basic structure of Resnet-50.

In order to train the deep network effectively and solve the problem of network degradation, bypass mechanism is introduced to resnet. The residual mapping replaces the original mapping in network. For networks with different depths, two structures are designed: One is used for ResNet34, another is for resnet50/101/152 and so on. In resnet, a compulsory module is called a residual block. Its structure is shown in Fig 3.

3D resnet(R3D) generates when 3D convolution operations are introduced into the resent network. It can be used for process of video to obtain spatiotemporal information. The structure of 3Dresnet-34 and 50. (R3D-34, R3D-50) are shown in Table.1.

**Table.1**. The parameter structure of Resnet-34/50

| Layer name | Output size | R3D-34 | rMC-34 |
|---|---|---|---|
| conv1 | L×56×56 | 3×7×7,64,stride 1×2×2 | |
| conv2_x | L×56×56 | $\begin{bmatrix} 3\times3\times3,64 \\ 3\times3\times3,64 \end{bmatrix} \times 3$ | $\begin{bmatrix} 1\times1\times1,64 \\ 3\times3\times3,64 \\ 1\times1\times1,256 \end{bmatrix} \times 3$ |
| conv3_x | L/2×28×28 | $\begin{bmatrix} 3\times3\times3,128 \\ 3\times3\times3,128 \end{bmatrix} \times 3$ | $\begin{bmatrix} 1\times1\times1,128 \\ 3\times3\times3,128 \\ 1\times1\times1,512 \end{bmatrix} \times 4$ |
| conv4_x | L/4×14×14 | $\begin{bmatrix} 3\times3\times3,256 \\ 3\times3\times3,256 \end{bmatrix} \times 3$ | $\begin{bmatrix} 1\times1\times1,256 \\ 3\times3\times3,256 \\ 1\times1\times1,1024 \end{bmatrix} \times 6$ |
| conv5_x | L/8×7×7 | $\begin{bmatrix} 3\times3\times3,512 \\ 3\times3\times3,512 \end{bmatrix} \times 3$ | $\begin{bmatrix} 1\times1\times1,512 \\ 3\times3\times3,512 \\ 1\times1\times1,2048 \end{bmatrix} \times 3$ |
| | 1×1×1 | Spatiotemporal pooling, Fc layer with softmax | |

The R2D and R3D are used widely in the feature extraction networks because of the excellent properties in deep network. In our model, R2D is used to extract the spatial feature while R3D is used to extract temporal feature. They are combined to realize the fusion of spatiotemporal feature before the classification of actions.

**3. Action recognition for specific target based on mixed convolutional Resnet and RPN**

Based on the mixed convolution Resnet, RPN and C3D are coupled in the feature extraction layer in this paper. The network extracts 2D image features in shallow layer and extracts temporal features in deep layer, realizing the parameter sharing of shallow features.

*3.1. Mixed convolutional Resnet*

It is very important to realize the separation of spatial and temporal operation of 3Dcnn. The mixed convolution network is also mentioned in [17] to be a contrast model to highlight the excellent effect of the decoupling 3Dresnet residual block. According to this theory, rMC-50 is designed in this paper, in which conv1-conv4x are 2D convolutional layers, conv5x is 3D convolutional layer. Its parameters and structure are shown in Fig.4. and Table 2.

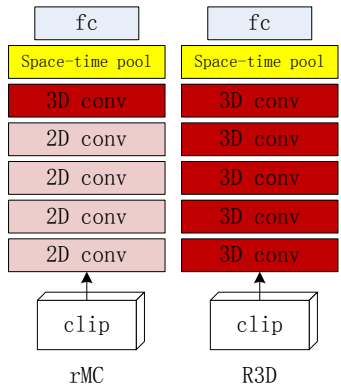

**Fig.4.** The structures of R3D and rMC

**Table.2.** Parameter structure of rMC and R3D

| Layer name | R3D-50 | | rMC-50 | |
|---|---|---|---|---|
| | output size | parameters | output size | parameters |
| conv1 | L×56×56 | 3×7×7,64,stride 1×2×2 | L×56×56 | 3×7×7,64,stride 1×2×2 |
| conv2_x | L×56×56 | [1×1×1,64; 3×3×3,64; 1×1×1,256] ×3 | L×28×28 | [1×1,64; 3×3,64; 1×1,256] ×3 |
| conv3_x | L/2×28×28 | [1×1×1,128; 3×3×3,128; 1×1×1,512] ×4 | L×14×14 | [1×1,128; 3×3,128; 1×1,512] ×4 |
| conv4_x | L/4×14×14 | [1×1×1,256; 3×3×3,256; 1×1×1,1024] ×6 | L×7×7 | [1×1,256; 3×3,256; 1×1,1024] ×6 |
| conv5_x | L/8×7×7 | [1×1×1,512; 3×3×3,512; 1×1×1,2048] ×3 | L/8×7×7 | [1×1×1,512; 3×3×3,512; 1×1×1,2048] ×3 |

The model is tested under the UCF101 dataset and compared with R3D-50 to verify the effectiveness of the mixed convolution operation. The results are shown in Fig.5.

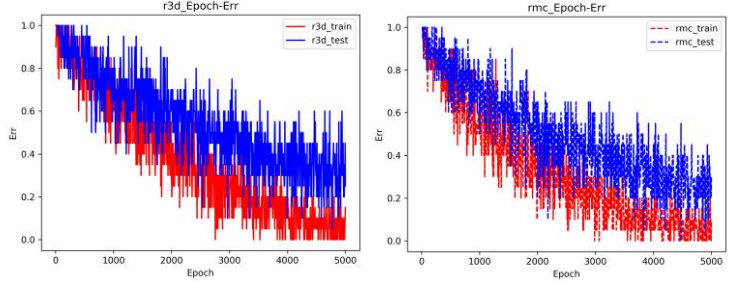

**Fig.5.** Error-epoch diagram for rMC and R3D

As shown in Figure 5, rMC-50 gets extremely good performance. When correct rate of train reaches 0.9-1, the correct rate of test reaches 0.7-0.8 in a batch in which there are 20 videos.
The correct rate of all test video reaches 71.07% when R3D reaches 66.65%. It shows that rMC-50 is effective to realize the action recognition. The way to separate two and three dimensional data is proved more effective than R3D.

*3.2. rMC-RPN*

According to section 3.1, rMC-50 network has the function of extracting spatial and temporal information which is similar to R3D. Based on it, RMC Network and R2D-50 network are coupled in network structure before conv5_x, in which the parameters are shard. the result of RPN for target detection is used to extract the feature maps on conv4-x to the crop pool layer .Further ,the feature maps are brought into the conv5_x of rMC for action recognition. Its network organization is shown in Figure 6.

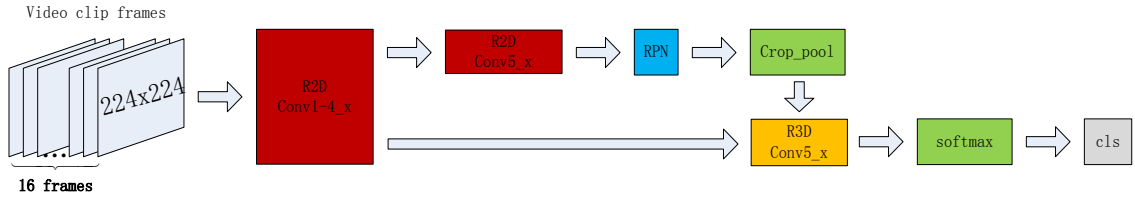

**Fig.6.** The structure of rMC based on RPN

*3.3. The losses of network*

As the structure of network of section 3.2 shown, the clip frames of video are processed to the same size(The size is $[224 \times 224]$) firstly. As the operation in this paper, feature maps with the size $[batch\_num \times L, 14, 14, 512]$ will be gotten after R2D con5_x. According to the detection position of RPN for each frame, the corresponding part of 2D conv4_x is brought into crop pool layer, which will be resized to $[batch\_num, L, 7, 7, 512]$. The result whose size is $[bathch\_num, act\_num]$ will be obtained after the R3D cov5_x and softmax layers. The total losses $loss_{total}$ consist of the classification loss of RPN $loss_{rpn\_cls}$, the regression loss of RPN and the classification loss of actions .

$$loss_{total} = loss_{rpn\_cls} + loss_{rpn\_reg} + loss_{act\_cls} \qquad (1)$$

$$loss_{rpn\_cls} = \frac{-\sum_{i}^{N_{cls}} scores_{rpn\_i} * tf(label_{rpn\_i})}{N_{cls}} \qquad (2)$$

$$loss_{rpn\_reg} = \lambda_1 \frac{\sum_{i}^{N_{reg}} \sum_{j}^{4} R(t_j^* - t_j)}{N_{reg}} \quad (3)$$

$$loss_{act\_cls} = -scores_{act} * tf(label_{act}) \quad (4)$$

where $N_{cls}$ is the number of valid anchor boxes in the task for classification of RPN. $scores_{rpn\_i}$ is the predicted score of anchor box i. $label_{rpn\_i}$ is the label of anchor box i in Eq.2. $i = [1,...,N_{cls}]$. $\lambda_1$ is the weights of $loss_{rpn\_reg}$ ($\lambda_1 = 3$). $N_{reg}$ is the number of valid anchor boxes for location of RPN, $i = [1,...,N_{reg}]$. $t_j$ is the transform parameters between ground truth boxes and valid anchor boxes,. $t_j^*$ is the transform parameters between prediction boxes and valid anchor boxes. $t_j^* \in [t_{xc}^*, t_{yc}^*, t_h^*, t_w^* t_w^*]$. R is the loss function of $Smooth_{L_1}$ in Eq.3. $scores_{act}$ is the predicted scores, $label_{act}$ is the label of video in Eq.4.

## 4. Experiments

*4.1. The introduction of experimental environment and data sets*
Hardware equipment: CPU: Rui long 2600; Video card: GTX1080Ti
Software Environment: Ubuntu16.04+pycharm carrying Tensorflow-GPU.
Training set: Our model is aimed at the action recognition for specific target, while the current action recognition dataset is not for specific target. There is also no detection data label in the data set. The specific gesture dataset is made by camera in this paper, as well as the action dataset with the interference target. In the gesture dataset, only the gesture movement changes, the share of hand in full picture share is very small. There are 10 kinds of actions and 60 videos in the data set, which are divided into right and left hand with 3 kinds of background. Except for the detection target, there exists inference targets in the body posture data sets. This date set consists of 10 segments of long video with the same actions. Data introduction is shown in the Figure.7.

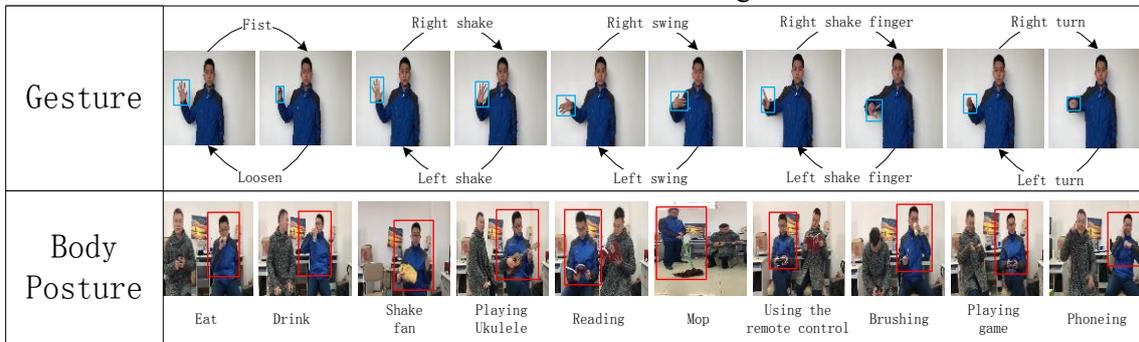

**Fig.7.** The description of data sets

*4.2. Experiment result of gesture data sets*

To train our model in gesture data set, 40 videos are put into network as train date set while other 20 videos are test data set. The model are trained 5000 times while each batch consisting 5 videos. The loss and error rate of each batch are recorded every 5 times. The losses and error ratio are shown in Fig.8 and a part of result of detection are shown in Fig.9.

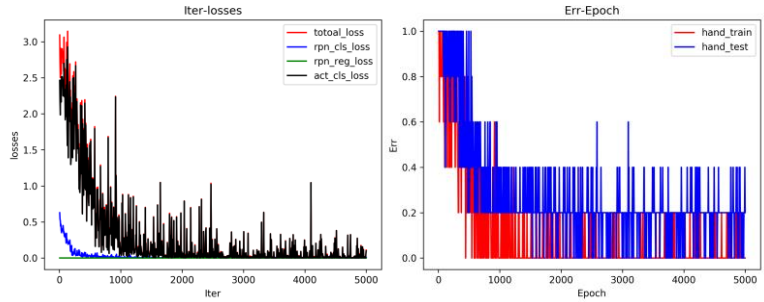
**Fig.8.** Iter-losses and Error-epoch of gesture data

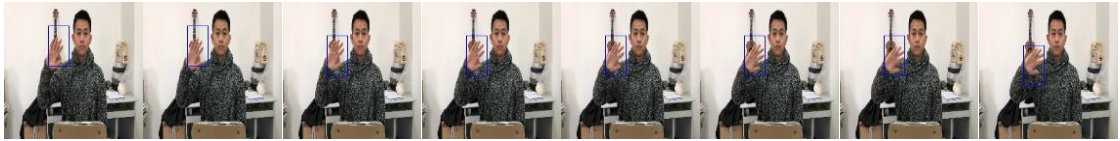
**Fig.9.** Result of detection for gesture

As shown in Fig.8, all losses gradually decline over time, and $loss_{act\_cls}$ is falling rapidly when $loss_{rpn\_reg}$ drops. It shows the action can be recognized when location of target is correct. All the losses will drop to a low level finally.

The error rate of each epoch is also shown in Fig.8. The training error can drop to 0 when test error drop to 0.2-0 eventually. The AP of detection reaches 44.81% and correct ratio of action recognition reaches the 65.3%.

This experiment of gesture data set shows this model is suitable for micro action recognition. Because the scale of anchors can be adjusted, the model can be adjusted for micro action recognition or the action of tiny object in the image. The model is useful for the gesture control and the action analysis of tiny object in videos.

*4.3. Experiment result of body posture data set*

To train our model in body posture data set, there are 10 long videos with same labels as data set. There is another person whose action is random except the valid target. Two thirds of the each video is produced as train sets while other part is test sets. The model are also trained 5000 times, and each batch consists 5 videos. The loss and error rate of each batch of body posture data are recorded every 5 times, which is shown in Fig.10. The result of detection is shown in Fig.11.

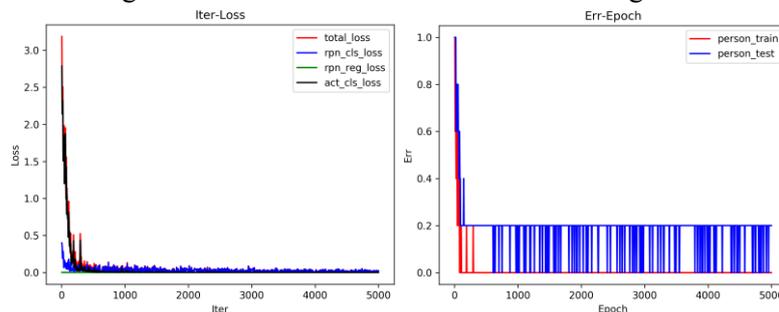
**Fig.10.** Iter-losses and Error-epoch of posture data

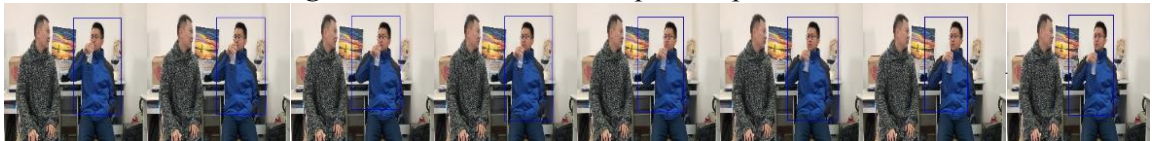
**Fig.11.** Result of detection for body posture

AS shown in Fig.10, the dropping for the losses of body posture data are similar to gesture's. But the losses drop faster than the hand's because the size of person is large suitable for anchors.

The error rate of each epoch is also shown in Fig.11. The training error can drop to 0 when test error drops to 0.2-0 quickly. In the last, the AP of detection reaches 42.75% and correct ratio of action recognition reaches the 64.8%. Our network can detect a video consuming 0.08s, which means the running speed can reaches 200 FPS.

This experiment of body posture data sets shows this model is effective for action recognition with interference target while the traditional method of action recognition can't solve this problem.

*4.4. Improvement of model*

According to the result of all the experiments, the AP of network is not very high because of the RPN's inaccuracies. The model can be improved when the correction of detect is higher.

So we improve our model according to the regression of proposal boxes in RCNN. A regression block is introduced between the crop pool and R3D conv5x layers in Fig.6. The new block is shown in Fig.12.

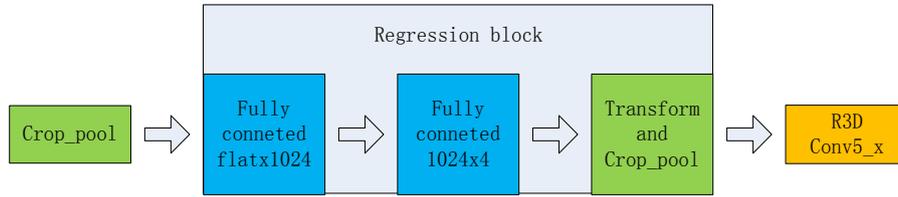

**Fig.12** The structure of regression block

As shown in Fig.12, the result of crop pool layer is flatted before put into the regression block. Then a group of new regression parameters is obtained after two fully connected layer. The number of hide point of the former is 1024, and the latter is 4.

To train the regression block, a new loss $loss_{roi\_reg}$ is added to the total loss.

$$loss_{total} = loss_{rpn\_cls} + loss_{rpn\_reg} + loss_{act\_cls} + loss_{roi\_reg} \qquad (5)$$

$$loss_{rpn\_reg} = \lambda_2 \sum_{j}^{4} R(tr_j^* - tr_j) \qquad (6)$$

where $\lambda_2$ is the weight of $loss_{roi\_reg}$ ($\lambda_2 = 1$). $tr_j$ is the transform parameters between ground truth boxes and proposal boxes from RPN, i.e., $tr_j \in [tr_{xc}, tr_{yc}, tr_h, tr_w]$. $tr_j^*$ is the transform parameters between prediction boxes and proposal boxes, i.e., $tr_j^* \in [tr_{xc}^*, tr_{yc}^*, tr_h^*, tr_w^*]$. R is the loss function of $Smooth_{L_1}$ in Eq. 6.

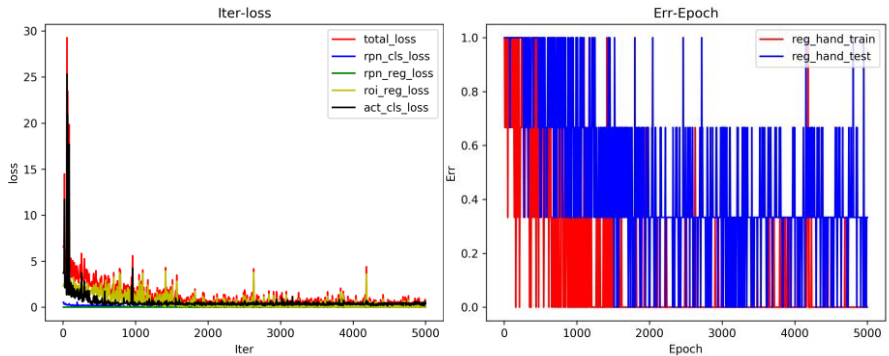

**Fig.13**. Iter-losses and Error-epoch of improved model

The model is trained in the gesture data with the same conditions as the previous model. Because of the increased number of parameters, the size of each batch decrease to 3. The result of improved model is shown in Fig.13 and Fig.14.

As shown in Fig.13, $loss_{roi\_reg}$ and $loss_{act\_cls}$ are high when the location of RPN is wrong. The model converges quickly as the location of RPN is correct and the adjustment of regression block.

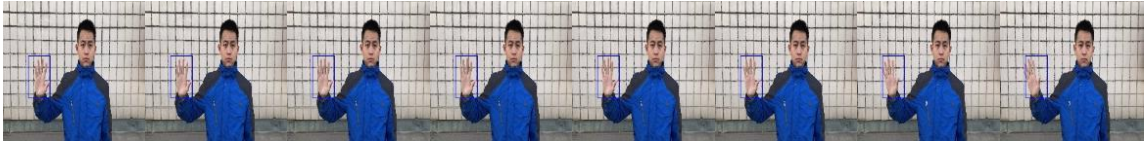

**Fig.14.** Result of detection of improved model

As shown in Fig.9 and Fig.14, the shape of hand are similar in these two actions, and the boxes of latter are tighter than the former. Test in gesture data set, the improved model's AP reaches 49.7% and accuracy of action recognition reach 70.7% while its running speed reaches 117 FPS which still performs very well The better performance of improved model shows our model has extremely powerful portability and potential for improvement. This model can be used as the infrastructure for action recognition of targets.

## 5. Conclusion

In this paper, based on rMC and RPN, the network is proposed to recognize action of specific target. Firstly, the effectiveness of rMC to recognize actions is tested in UCF10 and correct rate of it is higher than 3D Resnet(R3D). Through the composite of RPN and rMC in feature extraction layer, the model can be used to recognize action for specific target generates and its performance is tested in hand and body posture data sets. In experiments, the model achieves pretty good performance while operating at 200 FPS. Furthermore, regression block is introduced into our model to improve the effect. The ideal result shows the great potential of model.

**References**
[1] Girshick R , Donahue J , Darrell T , et al. Rich Feature Hierarchies for Accurate Object Detection and Semantic Segmentation[C]// 2014 IEEE Conference on Computer Vision and Pattern Recognition (CVPR). IEEE Computer Society, 2014.
[2] Girshick R. Fast R-CNN[J]. Computer Science, 2015.
[3] Ren S, He K, Girshick R, et al. Faster R-CNN: towards real-time object detection with region proposal networks[C]// International Conference on Neural Information Processing Systems. 2015.
[4] Redmon J, Divvala S, Girshick R, et al. You Only Look Once: Unified, Real-Time Object Detection[J]. 2015.
[5] Redmon J , Farhadi A . [IEEE 2017 IEEE Conference on Computer Vision and Pattern Recognition (CVPR) - Honolulu, HI (2017.7.21-2017.7.26)] 2017 IEEE Conference on Computer Vision and Pattern Recognition (CVPR) - YOLO9000: Better, Faster, Stronger[C]// IEEE Conference on Computer Vision & Pattern Recognition. IEEE, 2017:6517-6525.
[6] Redmon J , Farhadi A . YOLOv3: An Incremental Improvement[J]. 2018.
[7] Liu W, Anguelov D, Erhan D, et al. SSD: Single Shot MultiBox Detector[J]. 2015.
[8] Dai J, Li Y, He K, et al. R-FCN: Object Detection via Region-based Fully Convolutional Networks[J]. 2016.
[9] Simonyan K, Zisserman A. Two-Stream Convolutional Networks for Action Recognition in Videos[J]. 2014.
[10] Feichtenhofer C , Pinz A , Zisserman A . Convolutional Two-Stream Network Fusion for Video Action Recognition[J]. 2016.
[11] Bhattacharjee P , Das S . Two-Stream Convolutional Network with Multi-level Feature Fusion for Categorization of Human Action from Videos[J]. 2017.
[12] Tran D , Bourdev L , Fergus R , et al. Learning Spatiotemporal Features with 3D Convolutional Networks[J]. 2014.

[13] Jiang Z , Rozgic V , Adali S . Learning Spatiotemporal Features for Infrared Action Recognition with 3D Convolutional Neural Networks[J]. 2017.
[14] Ullah I , Petrosino A . Spatiotemporal Features Learning with 3DPyraNet[M]// Advanced Concepts for Intelligent Vision Systems. Springer International Publishing, 2016.
[15] Qiu Z, Yao T, Tao M. Learning Spatio-Temporal Representation with Pseudo-3D Residual Networks[J]. 2017.
[16] Chollet, François. Xception: Deep Learning with Depthwise Separable Convolutions[J]. 2016.
[17] Tran D , Wang H , Torresani L , et al. A Closer Look at Spatiotemporal Convolutions for Action Recognition[J]. 2017.
[18] He K , Zhang X , Ren S , et al. Deep Residual Learning for Image Recognition[J]. 2015.
[19] He K, Gkioxari G, Dollar P, et al. Mask R-CNN[J]. IEEE Transactions on Pattern Analysis & Machine Intelligence, 2017, PP(99):1-1.